\documentclass[11pt,twocolumn]{article}
\usepackage{pslatex, graphicx, amssymb, amsmath}
 \usepackage{url}
\def\abstract{
\typeout{Abstract}
 {\bf Abstract} 
}

\begin{document}
\title{An Unsupervised Anomaly Detection in Electricity Consumption Using Reinforcement Learning and Time Series Forest Based Framework}

\author{Jihan Ghanim \\ Department of Electrical and Computer Engineering, American University of Beirut,\\ Beirut, Lebanon
        \and Mariette Awad \\ Department of Electrical and Computer Engineering, American University of Beirut,\\ Beirut, Lebanon}

\maketitle
\begin{abstract}
Anomaly detection (AD) plays a crucial role in time series applications, primarily because time series data is employed across real-world scenarios. Detecting anomalies poses significant challenges since anomalies take diverse forms making them hard to pinpoint accurately. Previous research has explored different AD models, making specific assumptions with varying sensitivity toward particular anomaly types. To address this issue, we propose a novel model selection for unsupervised AD using a combination of time series forest (TSF) and reinforcement learning (RL) approaches that dynamically chooses an AD technique. Our approach allows for effective AD without explicitly depending on ground truth labels that are often scarce and expensive to obtain. Results from the real-time series dataset demonstrate that the proposed model selection approach outperforms all other AD models in terms of the F1 score metric. For the synthetic dataset, our proposed model surpasses all other AD models except for KNN, with an impressive F1 score of 0.989. The proposed model selection framework also exceeded the performance of GPT-4 when prompted to act as an anomaly detector on the synthetic dataset. Exploring different reward functions revealed that the original reward function in our proposed AD model selection approach yielded the best overall scores. We evaluated the performance of the six AD models on an additional three datasets, having global, local, and clustered anomalies respectively, showing that each AD model exhibited distinct performance depending on the type of anomalies. This emphasizes the significance of our proposed AD model selection framework, maintaining high performance across all datasets, and showcasing superior performance across different anomaly types.
\end{abstract}


\section{Introduction}
Anomaly detection (AD) finds widespread use in various applications \cite{zhang2022time}, including smart grids \cite{alkuwari2022anomaly}, Internet of Things (IoT) \cite{chatterjee2022iot}, medical systems \cite{khalil2022efficient}, financial fraud detection systems \cite{hilal2022financial}, and blockchain networks \cite{hassan2022anomaly}. With the prevalence of cloud infrastructure and IoT-enabled devices, there is a vast availability of time series data  \cite{singh2017cloud}. Consequently, AD for time series data is gaining attraction, particularly in data-driven applications \cite{jung2021time}. Anomalies, also called outliers, represent observations that deviate from the rest of the dataset \cite{barnett1994outliers} or exhibit deviations from most of the sequence's distributions or patterns \cite{ruff2021unifying}. For example, anomalies in power grid meter measurements may indicate potential cyber-attacks or malfunctions \cite{gunduz2018analysis, tehrani2019cyber}. Similarly, anomalies in time series power consumption patterns can lead to errors in power consumption predictions, resulting in underproduction or overproduction of electricity \cite{ghanim2022asymmetric}. In finance, outliers in time series can signify "illegal activities like fraud, risk, network intrusions, account takeover, and money laundering" \cite{hilal2022financial}. Thus, time series AD plays a crucial role in ensuring operational efficiency, reliability, and security in real-world applications \cite{jung2021time}.

However, AD often falls short of achieving expected performance in data-driven real-world systems. This is primarily due to the scarcity, cost, and impracticality of obtaining labeled data in large quantities \cite{jung2021time}. In real-world applications, a limited quantity of labeled anomalies can be provided through methods like active learning or identification by domain experts \cite{aissa2016semi,akcay2019ganomaly, han2022adbench}. Therefore, it is preferable to employ unsupervised AD techniques that do not rely on labeled data \cite{jung2021time}.

Furthermore, given the complexity of data in real-world systems, data abnormalities manifest in various forms \cite{chandola2009anomaly}. AD techniques that assume specific abnormalities may tend to rely on certain aspects more than others, making them sensitive to specific types of anomalies while potentially overlooking others. Consequently, there is a need for a universal AD model that outperforms other AD models on different types of data \cite{zhang2022time}. To address this, leveraging multiple AD models at different time steps can be beneficial when applied to a particular time series data. This can be achieved by designing an AD selection model that chooses the optimal AD model from a pool of candidate models at each time step \cite{chhabra2019method}. The predicted label (normal or anomalous) at each time step depends on the output of the currently selected anomaly detector. Reinforcement learning (RL), a machine learning domain that maximizes numerical rewards by mapping situations to actions at each time step, can be employed for this purpose \cite{sutton2018reinforcement}.

In a RL setting, an agent aims to learn the optimal decision-making policy to maximize the overall reward. RL has demonstrated success in various model selection applications, such as RL model selection for wind speed prediction \cite{kosana2022novel} and RL-based model combination for time series forecasting \cite{fu2022reinforcement}.

This study proposes an RL-based model selection approach for AD (RLAD) in time series datasets. To address the challenge of gathering labeled data which is both expensive and time-consuming \cite{karam2021progressive}, we focus on six unsupervised AD techniques that do not require labeled data for training. To implement the RL agent without explicit reliance on ground truth labels in its reward function, we utilize six time series forest (TSF) models. Each TSF is trained to classify a specific AD model's incorrect and correct predictions. Our proposed AD model selection framework utilizes the classifications from these six classifiers for the majority of data points and ground truth labels for the remaining data points \cite{deng2013time}. In other words, our approach requires a portion of ground truth labels instead of explicit reliance on all ground truth labels. We compare our RL-based model selection approach with an RL model selection that solely relies on ground truth labels in its reward function and another approach that relies only on the classifications of the TSFs in its reward function.

Our proposed RLAD framework yields promising results, outperforming all AD candidate models on the real dataset and all AD models except for KNN on the synthetic dataset. Additionally, we conduct a study on different reward scenarios, including adaptive rewards with varying exploration epsilons, in the context of our AD model selection approach. The results indicate that the original reward function with a decaying exploration rate produces the best scores in terms of precision, recall, and F1 score evaluation metrics. We also examine the impact of different types of anomalies on the performance of the AD techniques by testing them on three synthetic datasets: one with local anomalies, another with global anomalies, and a third with clustered anomalies. The performance of each AD model varies across different types of anomalies, whereas our proposed AD model selection approach consistently achieves high scores across global, local, and clustered datasets, respectively.

\section{Related Work}

Due to the emerging spread of time series problems \cite{lim2021time}, the field of AD has garnered substantial attention for the analysis of time series data \cite{braei2020anomaly}. It can be categorized into three main groups: unsupervised, semi-supervised, and supervised methods \cite{aggarwal2016outlier}. While supervised or semi-supervised methods can utilize previously identified abnormalities to label instances as normal or abnormal, the practicality of supervised techniques is limited by the need for a large amount of labeled data containing both normal and abnormal instances, which is often not feasible in real-world settings \cite{gornitz2013toward}. Semi-supervised AD techniques typically stem from supervised algorithms by incorporating a bias term to account for unlabeled data \cite{vapnik1999overview, sindhwani2005beyond}. In light of this, unsupervised AD categories, such as distance-based, statistical, ensemble-based, and reconstruction-based approaches, have recently gained favor. 

In this section,  we comprehensively understand these prominent categories of unsupervised AD techniques and their applicability in various scenarios. After exploring the strengths and limitations of these techniques, we discuss the recent advancements in this field, with the goal of tackling the existing AD challenges effectively.

\subsection{Distance-based Approaches}

Among the distance-based AD techniques,  \cite{knorr2000distance} considers a point (p) as an $(\pi,\varepsilon)$-anomaly if at most $(\pi)$\% of the entire data points are away from p by a distance less than $(\varepsilon)$. This was further improved in \cite{ramaswamy2000efficient} by proposing an outlier score which is obtained by calculating the distance to the $k^{th}$ nearest neighbor (KNN) of a data point. After that, the data point is classified as normal or anomalous based on a threshold. \cite{angiulli2002fast} suggested a better variant with a KNN-based method where the outlier score represents the aggregate distance of the data point (p) to the k nearest neighbors of p. Another distance-based AD algorithm, the Local Outlier Factor (LOF), considered the ratio of the average of the k nearest neighbors densities to the data point density itself \cite{breunig2000lof}. While LOF and KNN have both been shown to outperform state-of-the-art (SOTA) AD techniques, they yet failed to preserve a high performance with large high-dimensional and seasonal datasets \cite{campos2016evaluation}.

\subsection{Statistical Approaches}

 Numerous seminal probabilistic AD methods compute the anomaly score relying on the estimates of the marginal likelihood ${\mathbb{P}_\theta(X)}$ derived from the data generation model \cite{xiong2011direct,li2009dynammo,gornitz2015hidden}. One of the probabilistic techniques that fits a number of Gaussians on the data is Gaussian Mixture Model (GMM). To estimate the models' parameters, the expectation-maximization algorithm (EM) is utilized \cite{dawid1979maximum}. Support vector machine (SVM) has been extensively used for novelty detection tasks \cite{efficientawad}. One-class SVM (OSVM) is a support-vector-based approach that assumes a probability density (like GMMs fitting) to learn a hyper-plane that separates the low-density region from the region with a high-density of data \cite{manevitz2001one, scholkopf2001estimating}. One limitation of generative models lies in the complexity of obtaining marginal likelihoods, which necessitates computing integrals with exceedingly high dimensions. ${\ \mathbb{P}_\theta(X) = \int \mathbb{P}_\theta(z) \mathbb{P}_\theta(X \mid z) \, dz}$ \cite{jung2021time}.
More recent non-parametric statistical methods for AD are Empirical Cumulative Distribution for Outlier Detection (ECOD) \cite{li2022ecod} and Copula-Based Outlier Detection (COPOD) \cite{li2020copod}. The ECOD approach involves computing an empirical cumulative distribution to predict tail probabilities for individual data points, which are then used to calculate the outlier score. In contrast, COPOD utilizes an empirical copula to estimate tail probabilities for the entire set of data instances, enabling the computation of the outlier score. Although these novel techniques provide valuable contributions, their scalability is constrained by their dependence on the data's shape \cite{kharitonov2022comparative}.

\subsection{Ensemble-based Approaches}

In \cite{xu2019automatic}, an ensemble-based AD technique considered Local Outlier Factor (LOF) with multiple hyper-parameter sets and combined the results to obtain outlier scores. Another ensemble-based technique, known as Isolation Forest (IForest),  was proposed in \cite{liu2008isolation}. IForest consists of a forest with random binary trees in such a way that outliers are relatively near the tree's root (within a shallow depth of the tree). The underlying principle of IForest is based on the notion that outliers are more prone to isolation. 

\subsection{Reconstruction-based Approaches}

Several contemporary AD algorithms capitalize on using synthetic data reconstruction. These algorithms are based on the insight that outliers, when projected into a lower-dimensional space, tend to lose information, leading to less efficient reconstruction and an increased reconstruction error for anomalous instances  \cite{jung2021time}. Principal Component Analysis (PCA) is a popular approach for data reconstruction that permits linear reconstruction only \cite{rousseeuw2018anomaly}. The non-linear variant of PCA, known as Kernel PCA, applies the kernel trick to map the data into a feature space \cite{hoffmann2007kernel}. Deep learning-based AD algorithms, such as those employing Recurrent Neural Networks (RNN), have garnered significant attention, especially in time series applications \cite{malhotra2016lstm, su2019robust}. However, these techniques can be computationally expensive and time-consuming \cite{hochreiter1997long, el2021optimized}. A more recent approach is Unsupervised AD on Multivariate Time Series (USAD) \cite{audibert2020usad}. It relies on an encoder-decoder pair that is trained in an adversarial manner. The reconstruction error in the testing phase represents the anomaly score.

\subsection{Large Language Models-based Approaches}

Large Language Models (LLMs) have lately shown remarkable success in the AD task. LLMs are capable of processing and interpreting huge amounts of data having temporal dependencies which provide more complicated and precise AD compared to traditional statistical methods. Motivated by the fact that anomaly detection of time series data resembles text classification and that BERT \cite{gonzalez2020comparing} has proven effective in handling text classification, Dang et al. \cite{dang2020time} suggested a time series anomaly detection technique relying on BERT model. Their model has the same architecture as the BERT model, however, the main distinctions are the input representation and additional output layer. Simulation results reveal that this technique only requires a small amount of labeled data for training the BERT model to achieve better results than the SOTA methods. Yet, the model scores on the KPI dataset are still not high enough recording an F1 score of 0.634. Recently, Dong et al. \cite{dong2024can} studied the performance of zero-shot learning of LLMs in time series AD problems and their explanatory abilities. They also introduced a synthesized dataset to fine-tune LLMs and improve their performance in time series AD. Their work states that LLMs exhibit encouraging potential for time series AD, while customized prompts and instructions remain necessary. Despite that GPT-4 showed good results with minimal instructions, the present absence of public fine-tuning capabilities on GPT-4 stopped them from investigating whether fine-tuning on GPT-4 could record SOTA performance.

Leveraging LLMs for AD holds significant promise but still faces several critical challenges. One main concern is their dependence on historical datasets, which can be a matter of concern related to data availability, quality, and model bias. Moreover, these models often have trouble with generalizability, making it hard for them to use learned patterns with novel scenarios. LLMs may also generate inaccurate or misleading outputs under certain conditions, due to hallucinations, which question their reliability. At last, the computational efficiency of these large models imposes a barrier, as their resource-intensive nature can limit scalability and accessibility \cite{su2024large}.

\subsection{Model Selection for AD}

Most work in AD has attempted to build novel and enhanced models to perform outlier detection on distinct data types. However, very limited work exists on the AD model selection task. Research work in \cite{li2021autood} proposed an automated outlier detection method that applied neural architecture search to find the best neural network model. The limitation of this technique is that it is restricted to auto-encoder-based AD methods and relies on annotated data for evaluation. Another set of automated-based AD model selection techniques are TODS \cite{lai2021tods} and PyODDS \cite{li2020pyodds} which also rely on labeled data. A more recent automated model selection for unsupervised outlier detection proposed a meta-learning approach without needing ground truth labels in the testing phase \cite{zhao2021automatic}. Yet, their approach requires ground truth labels in the meta-train datasets. These model selection methods tend to choose a single AD model after evaluation. However, a recent model selection approach applied to the Secure Water Treatment (SWaT) dataset chooses dynamically an unsupervised AD technique among five different AD models using RL \cite{zhang2022time}. However, their AD pool of models does not contain a distance-based AD model. Additionally, their RL framework does not consider the anomalous dataset in its state space. Another drawback is that their RL approach relies explicitly on ground truth labels in its reward function.
Due to the difficulty of obtaining a huge amount of labeled data in real-world applications, we propose a dynamic model selection for time series AD without excessively relying on ground truth labels. Our approach uses a time series forest and RL-based framework relying on a limited amount of ground truth labels to select an AD algorithm from six unsupervised AD techniques at each time step. Specifically, we will consider six various and popular AD algorithms from the above-mentioned categories: KNN, COPOD, ECOD, OSVM, IFOREST, and USAD.

\section{Methodology}
\subsection{Unsupervised AD for Time Series}

\begin{figure*}[t]
    \centering
\includegraphics[scale=0.6]{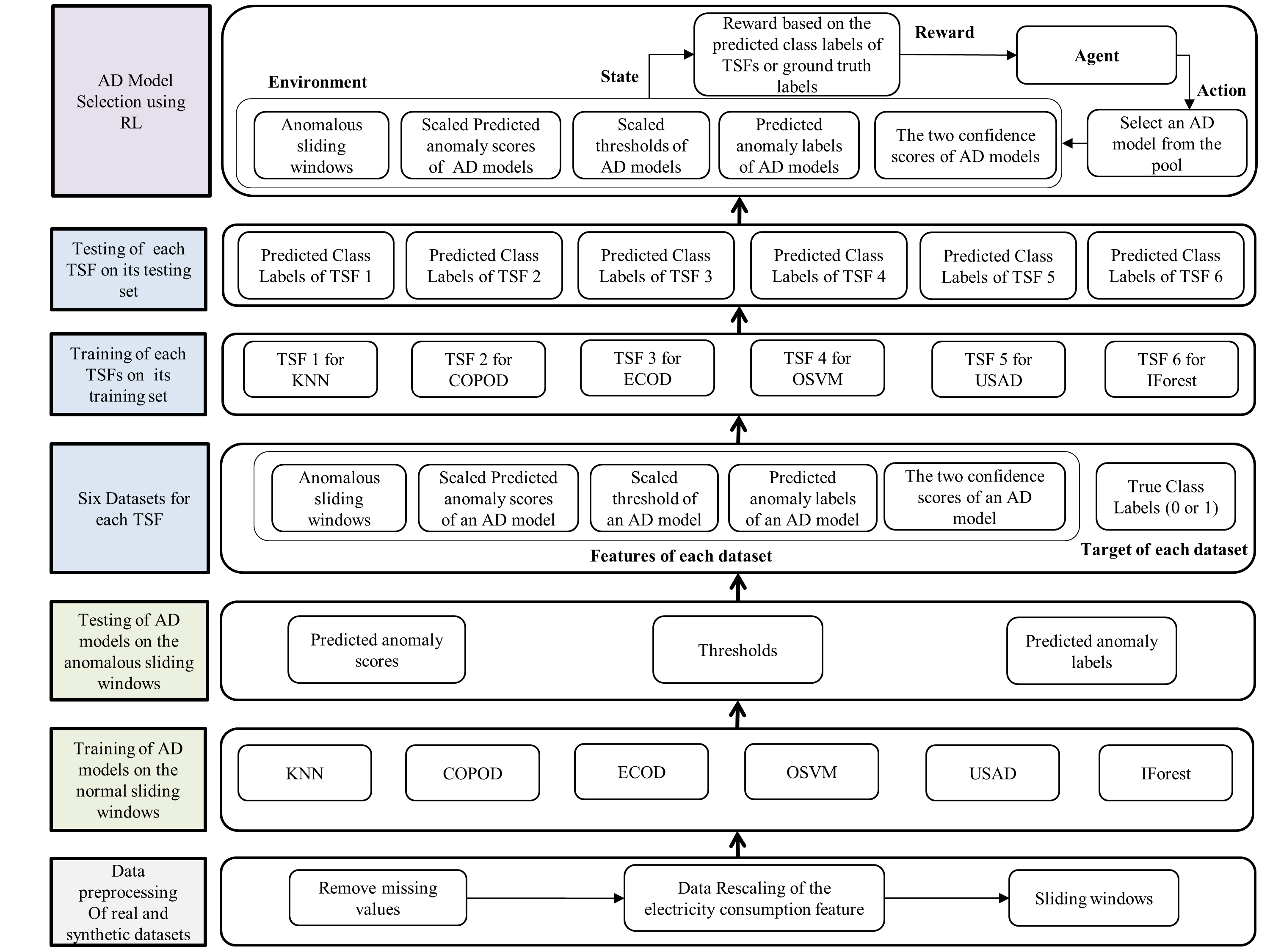}
    \caption{Illustration of the Proposed Framework}
    \label{system_framewok}
\end{figure*}

This research focuses on building a model selection for time series unsupervised AD. Figure \ref{system_framewok} illustrates our proposed RLAD model selection. Six distinct unsupervised AD techniques are considered in our framework; KNN \cite{angiulli2002fast}, COPOD \cite{li2020copod}, ECOD \cite{li2022ecod}, OSVM \cite{manevitz2001one}, IForest \cite{liu2008isolation}, and USAD \cite{audibert2020usad}. First, the training set D$_{normal}$, containing a time series of only normal data points, is separately fed to each unsupervised AD model. Every AD model is separately trained on the training set D$_{normal}$. Then, every trained model is tested on the testing set D$_{anomaly}$ having anomalous data points. Each model will output a set of anomaly scores; a score for each instance of the testing set D$_{anomaly}$. For every AD model, an anomaly threshold is empirically computed. Depending on the produced anomaly scores and thresholds, each AD model yields a predicted anomaly label for every instance of D$_{anomaly}$.

\subsection{The Confidence Scores}
Two confidence scores, initially used by \cite{zhang2022time}, were considered in our proposed model selection framework for AD techniques. The first confidence score is derived from the concept that the greater the anomaly score, the more likely the corresponding instance is anomalous. 
AD models generate anomaly scores when tested on new data. The higher the predicted anomaly score exceeds the model's threshold, the more likely it is that the tested data point is an anomaly. Therefore, the extent to which the anomaly score surpasses its threshold serves as a reasonable indicator of the prediction's reliability for the given AD model. The Distance-to-Threshold Confidence Score is equivalent to the following: 

\begin{equation}
\begin{split}
\frac{\text{Anomaly \ score} - \text{Threshold}}{\max(\text{Anomaly \ score} )- \min(\text{Anomaly \ score})}
\end{split}
\end{equation}

It is worth mentioning that the anomaly scores are scaled using the min-max scale to a range between zero and one before computing the Distance-to-Threshold Confidence score.
Inspired by the principle of majority voting in ensemble learning, the Prediction Consensus Confidence Score indicates that as more AD models in the pool produce the same prediction (predicted label), the likelihood of the current prediction being correct increases. Precisely, the Prediction Consensus Confidence Score is determined as follows:
\begin{equation}
\begin{split}
\frac{\text{Number \ of \ models\ predicted \ the \ same \ label}}{\text{Total \ number\ of \ models}}
\end{split}
\end{equation}

\subsection{Evaluation Metrics}
We annotated the normal data as “negative” and the anomalous one as “positive”. To evaluate the performance of our proposed approach, we apply the well-known precision, recall, and F1score metrics.

\begin{align}
\text{Precision} & = \frac{TP}{TP + FP} \label{eq:precision}
\end{align}

\begin{align}
\text{Recall} & = \frac{TP}{TP + FN} \label{eq:recall} 
\end{align}

\begin{align}
\text{F1score} & = \frac{2 \times \text{precision} \times \text{recall}}{\text{precision} + \text{recall}} \label{eq:f1score}
\end{align}
 
where the elements of the confusion matrix are: 

\textbf{True Positive (TP)}: represents the case where the anomalous instance is correctly predicted.

\textbf{False Positive (FP)}: stands for the case where the anomalous instance is predicted as a normal instance.

\textbf{True Negative (TN)}: refers to the case where the normal instance is correctly predicted as a normal instance. 

\textbf{False Negative (FN)}: refers to the case where the normal instance was predicted as an anomalous one. 

\subsection{Time Series Classification Using TSF}	
Six TSFs are implemented; a TSF for the predicted anomaly labels of KNN, COPOD, ECOD, OSVM, IForest, and USAD, respectively. Each TSF classifies the predicted anomaly labels of a specific trained AD model (from the previous step) into two classes, C=\{0,1\} where C=0 indicates wrongly predicted anomaly labels and C=1 indicates correctly predicted anomaly labels. In other words, if the anomaly-predicted label of a certain AD model is equivalent to the ground truth label, then it is classified as 1 otherwise it is classified as 0. 

Since each TSF corresponds to a specific AD model, then each TSF is applied to a different dataset containing the anomaly predictions of its corresponding AD model. The dataset of each TSF consists of a set of features and a target. The features of each dataset represent the anomalous instances D$_{anomaly}$, along with the corresponding predicted anomaly scores (after rescaling them) obtained from the corresponding AD model, the empirical scaled threshold of the corresponding AD model, the predicted anomaly labels of the corresponding AD model, and the corresponding two confidence scores. The target of each dataset consists of the true class labels (0 or 1) belonging to its corresponding AD model. The target in each dataset of the TSFs is obtained by comparing the predicted anomaly label of its corresponding AD model to the ground truth label.

During training, each of the TSFs is trained on its corresponding training set where 20\% of its corresponding dataset is considered for training. In the testing phase, each TSF is tested on its corresponding testing set, having the remaining 80\% of its corresponding dataset. Each of the six TSF classifiers predicts the class labels (0 or 1). 
The target in the training set of each TSF is obtained by comparing the anomaly-predicted label of its corresponding AD model to the ground truth label. Thus, 20\% of the anomalous instances and their ground truth labels are required to train each TSF classifier. The set of anomalous instances and ground truth labels used for all six datasets should not exceed 30\% of the anomalous dataset to ensure that a limited amount of ground truth labels is only needed.

\subsection{AD Model Selection Using  Deep Q Network (DQN)}

To tackle our model selection challenge AD, we adopt an RL framework. RL is a type of machine learning that involves an agent learning to make decisions by interacting with an environment. The agent receives feedback in the form of rewards or penalties based on its actions and uses this information to optimize its decision-making strategy over time. RL is particularly suitable for problems where the agent must learn from trial and error to maximize cumulative rewards, such as in robotics, game playing, and autonomous systems. RL can thus be conceptualized as a Markov Decision Process (MDP) where the RL agent maximizes the total expected return by learning the decision policy. MDP contains five major elements; an action-space consisting of the actions set, a state-space containing the states set, a transition matrix representing the probability of state-transition, a reward function, and a discount factor that refers to a value between 0 and 1 that is used in calculating the reward. The future return in an MDP consists of the immediate reward along with the discounted future rewards. The policy represents a probability distribution that maps the present state to the possibility of choosing a particular action.

In this work, the DQN algorithm of RL is applied. DQN is a form of RL that utilizes deep neural networks to approximate the action-value function, known as Q-function, in a high-dimensional state-action space. It combines the power of deep learning with Q-learning to enable the efficient handling of complex environments. The agent uses experience replay, storing past experiences in a memory buffer, and updates the neural network using batches of experiences, breaking the correlation between consecutive experiences. This technique allows for more stable and effective learning, enabling the agent to make informed decisions by selecting actions that yield the maximum expected future rewards.  In our MDP modeling, the DQN agent was trained to learn a policy to select the suitable AD model among the pool at each time step. At each time step, the agent was rewarded based on its action (selection).

Given the context of a time series application, the transitions between states adhere to the chronological order of the time series dataset. As a result, the state transition is considered deterministic, eliminating the necessity to incorporate a transition matrix. Moreover, we set the discount factor to one to appropriately handle the time series nature of the problem. The remaining components of the MDP  are described below;

\textbf{State-space}: The entire testing set D$_{anomaly}$, the scaled predicted anomaly scores from all the considered AD models, the obtained scaled empirical thresholds, and the predicted anomaly labels for these six AD models, as well as the two confidence scores of these AD models.

\textbf{Action-space}: The action-space is a discrete action-space. The DQN agent can select one AD model from the pool of the six AD models at each time step. Thus, the size of the action-space is the same as the number of candidate AD models in the pool.

\textbf{Reward Function (R)}: The reward function is obtained by comparing the predicted anomaly label obtained from the selected AD model in the pool of models to the predicted class labels obtained the corresponding TSF in most of the testing set $D_{anomaly}$ (80\% of $D_{anomaly}$). This represents the class-based portion of the reward. In the remaining 20\%, the reward function is obtained by comparing the predicted anomaly label obtained from the selected AD model in the pool of models and its corresponding ground truth labels. This refers to the ground truth-based portion of the reward. It should be mentioned that the same 20\% of ground truth labels used during the training of the TSFs, are used here in the reward function.

The reward function of equation \ref{eq3} is used as the original reward, where a TP is given a higher reward compared to a TN reward ($r_{TP} > r_{TN}$). A TP is an instance where the agent selects an AD model that correctly predicts an anomaly.  However, TN means that the AD model selected by the agent at this time step correctly predicts this data point as normal. Since the majority of instances are usually normal while the anomalous instances are rare and we are interested in detecting anomalies, thus the reward for true positive should be set to be greater than the TN reward $r_{TP}>r_{TN}$. On the other hand, a higher penalty should be applied to FN compared to a FP. A FN represents a case where the RL agent selects an AD model that predicted this data point to be normal when this point is actually anomalous. A FP refers to a case where the agent chooses an AD model that predicts this instance as anomalous when it is, in reality, normal. In real-world applications, neglecting anomalies (FN case) is more detrimental than giving false alarms (FP case), so $|{r_{FN}}|>|{r_{FP}}|$. As such, we reflected it in equation \ref{eq3} a reward function that consists of r$_{TN}$ that is set to 0.5, r$_{TP}$ is set to 1, r$_{FN}$ is set to -1.5, and r$_{FN}$ is set to -3. 

\begin{equation}
R = \left\{
\begin{array}{ll}
r_{TN} = 0.5 & \text{TN} \\
r_{FN} = -3 & \text{FN}  \\
r_{TP} = 1 & \text{TP}   \\
r_{FP} = -1.5 & \text{FP}
\end{array}
\right.
\label{eq3}
\end{equation}

\section{Dataset Description and Experimental Settings}
\subsection{Datasets Description}
In this work, we considered two datasets; one real and one synthetic dataset. Each of these datasets consists of the hourly time-steps, the electricity consumption, and the ground truth labels. 
The real dataset is an annotated subset of the Large-scale Energy AD dataset (LEAD1.0) \cite{gulati2022lead1}, and it is made up of hourly meter readings obtained from 1,413 smart electricity meters over one year and gathered from sixteen distinct sites around the world. The LEAD1.0 dataset contains point and sequential (collective) anomaly types. For the sake of this work, we used a subset of LEAD1.0 containing hourly meter readings from twenty-two smart electricity meters. We split this real data into a normal dataset consisting of only normal instances and an anomalous dataset with normal and anomalous instances. The normal dataset consisting of hourly electricity meter readings (electricity consumptions) has a total of 18696 time-steps. However, the anomalous hourly electricity meter readings dataset has readings of 20424 time-steps with around 5.7\% of anomalies and their corresponding ground truth labels.

The second dataset considered in this work is a synthetic dataset in which anomalies are injected randomly into the dataset. We used the time series of hourly electricity consumption data from the residential areas of Germany provided by ENTSO-E Transparency, a European electricity data platform \cite{opsd}. Similarly, this dataset was divided into normal and anomalous datasets. The normal dataset has hourly power-consumption data from the year 2015 until 2018 making up a total of 35064 time-steps. The anomalous dataset consists of 1008 hourly power-consumption data points of 2019 with 5\% of randomly injected anomalies and the corresponding ground truth labels.

\subsection{Data Preprocessing} Initially, data preprocessing was performed on the datasets. First, any missing values in the data were removed. Then, the electricity consumption data points in the normal and anomalous datasets (synthetic and real datasets) were rescaled. In the synthetic normal and anomalous datasets, we applied min-max scaling to scale the electricity consumption data into values between zero and one. However, in the real datasets, we applied logarithmic scaling similar to the original paper of LEAD1.0 \cite{gulati2022lead1}.  After that, each normal and anomalous dataset was transformed into sliding windows of size six with a step size of one. The chosen size of the sliding window and the step size were found the most suitable after heuristically trying sliding sizes six, twelve, and twenty-four with different step sizes of one, two, three, and zero overlaps.

\subsection{Settings of the AD Candidate Models in the Pool}
We considered six different AD models in our pool of models, which are KNN \cite{angiulli2002fast}, COPOD \cite{li2020copod}, ECOD \cite{li2022ecod}, OSVM \cite{manevitz2001one}, IForest \cite{liu2008isolation}, and USAD \cite{audibert2020usad}. OSVM and IForest are implemented from the Scikit-Learn library \cite{pedregosa2011scikit}.  COPOD, ECOD, and KNN are imported from the PyOD toolbox \cite{zhao2019pyod}. As for the USAD model, it was implemented from the GitHub repository of USAD \cite{usad}.

For each of KNN, OSVM, and IForest, we performed hyperparameter optimization using Hyperopt library \cite{bergstra2015hyperopt} to tune their hyperparameters and select the best configuration of hyperparameters. The list of hyperparameter choices for each of KNN, OSVM, USAD, and IForest AD models are available in the Appendix. COPOD and ECOD do not require hyperparameter tuning since they are deterministic and have no hyperparameters. As for USAD, it has two hyperparameters, alpha and beta, such that their sum must be equal to one. They are utilized for parameterizing the trade-off between true and false positives, which varies the model's sensitivity. We used the original implementation of USAD having neutral detection sensitivity where alpha equals beta equals 0.5 to avoid having a sensitive model biased towards true positives or false positives.

A specific criterion is necessary to compute the empirical threshold for each AD model. Given that we know the percentage of anomalies in both datasets, we set the criterion to be equal to this percentage (5\%).

To elaborate further, for each AD model, if a data point's predicted anomaly score ranks among the highest 5\%,   
within all the predicted anomaly scores generated by that model, the data point is classified as an anomaly.

\subsection{Settings of the Proposed RL Framework for AD Model Selection}

In the implementation of the RL framework, we used the DQN model imported from Stable-Baseline3. Stable-Baseline3 is a RL toolbox based on PyTorch \cite{raffin2021stable}. The default settings of DQN provided by \cite{raffin2021stable}, having a decaying exploration epsilon, were utilized. However, the exploration fraction of the decaying exploration epsilon was set to 0.7 instead of the default value of 0.1 to allow the DQN agent to explore more, and the seed number was set to 1. 
 In the synthetic anomalous dataset case, the RL agent was trained for 600,000 time steps. The agent was trained for 3,000,000 time steps in the real anomalous dataset case.  This is because the real dataset is larger than the synthetic one. During the evaluation phase of the DQN agent, we set the agent’s actions to be deterministic.

\section{Experiments and Results}
 
\subsection{AD Using the Candidate AD Models in the Pool}
To perform AD using the six unsupervised AD techniques, each  AD  model was trained on the normal sliding windows having normal electricity consumption instances. After that, each of these pre-trained AD models was tested on the sliding windows of the anomalous electricity consumption instances. The predicted anomaly label (anomalous or normal instance) was compared to the ground truth labels.

The results of the unsupervised AD models on the synthetic and real datasets are presented in Tables \ref{table:1} and \ref{table:2}, respectively. In the synthetic data, KNN achieved the highest performance having precision, recall, and F1score of 1. USAD and IForest models also achieved high performance, recording a score of 0.925 and 0.902, respectively in all the evaluation metrics. OSVM scored a lower performance having a value of 0.624 in all metrics. COPOD and ECOD yielded the lowest values of 0.361 and 0.414 in the evaluation metrics. 

As shown in Table \ref{table:2} for the real data, KNN also recorded the highest of 0.702 in all the evaluation metrics. The results showed that the KNN model outperformed the other AD models' performance in both datasets. IForest model came next with a 0.579 precision, recall and F1score. Following this, COPOD and ECOD models recorded a value of 0.532 in all metrics. However, OSVM and USAD both failed to detect anomalies.

Thus, KNN seemed to preserve the highest performance in both datasets, however, it recorded lower results in the real data. The remaining AD models showed a different performance on different datasets. These results further illustrate the need for a dynamic AD model selection that can guarantee accurate AD in time series applications, especially with the absence of large quantities of ground truth labels in real-world applications.

\begin{table}[t]
\begin{center}       

    \begin{tabular}{ |c|c|c|c| } 
                \hline
                    \textbf{AD Model} & \textbf{Precision} & \textbf{Recall} & \textbf{F1score} \\  
                \hline
                    COPOD & 0.361 & 0.361 & 0.361 \\
                \hline
                    ECOD & 0.414 & 0.414 & 0.414 \\
                \hline
                    KNN	& 1 & 1	& 1 \\
                \hline
                   OSVM & 0.624 & 0.624 & 0.624 \\
                \hline
                    IForest & 0.902	& 0.902 & 0.902 \\
                \hline
                   USAD & 0.925 & 0.925 & 0.925 \\
                \hline
                    \textbf{Proposed} & \textbf{1} & \textbf{0.977} & \textbf{0.989}  \\ 
                \hline
    \end{tabular}
            \caption{Performance of AD Candidate models on the synthetic anomalous dataset.}
        \label{table:1}
\end{center}
\end{table}

\begin{table}[t]
\begin{center}       

    \begin{tabular}{ |c|c|c|c| } 
                \hline
                    \textbf{AD Model} & \textbf{Precision} & \textbf{Recall} & \textbf{F1score} \\ 
                \hline
                    COPOD & 0.532 & 0.532 & 0.532 \\
                \hline
                    ECOD & 0.532 & 0.532 & 0.532 \\
                \hline
                    KNN	& 0.702	& 0.702 & 0.702 \\
                \hline
                   OSVM & 0.003 & 0.003 & 0.003 \\
                \hline
                   IForest & 0.579	& 0.579	& 0.579 \\
                \hline
                   USAD & 0.003 & 0.003 & 0.003 \\
                \hline
                    \textbf{Proposed} & \textbf{0.977} & \textbf{0.579} & \textbf{0.727}  \\
                \hline
    \end{tabular}
            \caption{Performance of AD Candidate models on the real anomalous dataset.}
        \label{table:2}
\end{center}
\end{table}

\subsection{Time Series Classification Using TSF}

To avoid relying on large quantities of ground truth labels that are not usually available in real-world applications, TSFs were used to classify the predicted anomaly labels into correct or wrong labels. Six TSFs were implemented corresponding to the six AD models. Each TSF was fed with a different dataset corresponding to the AD model.

The results of each of the TSFs on their corresponding testing set are shown in Tables 3 and 4 using anomalous sliding windows from the synthetic and real datasets respectively. It is worth mentioning that since KNN when tested on the synthetic anomalous data achieved a superior performance of 1, as shown in Table \ref{table:1}, in all precision, recall, and F1score metrics, then no need to apply a TSF to classify the predictions of KNN.  Table \ref{table:3} illustrates that all TSFs achieved remarkable performance of F1scores ranging between 0.987 and 0.997 in the case of anomalous synthetic instances. Similarly, in Table \ref{table:4}, in the case of the anomalous real dataset, they recorded F1scores ranging between 0.913 and 0.979.

\begin{table}[h]
\begin{center}       

    \begin{tabular}{|p{2.5cm}|c|c|c|c| } 
                \hline
                    \textbf{TSF based on predictions from} & \textbf{Precision} & \textbf{Recall} & \textbf{F1score} \\ 
                \hline
                    COPOD & 0.99 &	0.993 & 0.991 \\
                \hline
                    ECOD & 0.983 &	0.997 & 0.99 \\
                \hline
                    KNN	& -	& - & - \\
                \hline
                   OSVM & 0.989 &	0.996 &	0.993\\
                \hline
                   IForest & 0.9738 &	1 & 0.987 \\
                \hline
                   USAD & 0.998 & 0.996 &	0.997 \\
                \hline

    \end{tabular}
            \caption{Results of each of the TSFs on their corresponding testing set having anomalous sliding windows from the synthetic dataset.}
        \label{table:3}
\end{center}
\end{table}

\begin{table}[h]
\begin{center}       

    \begin{tabular}{|p{2.5cm}|c|c|c|c| } 
                \hline
                    \textbf{TSF based on predictions from} & \textbf{Precision} & \textbf{Recall} & \textbf{F1score} \\ 
                \hline
                    COPOD & 0.978 &	0.956 &	0.9667 \\
                \hline
                    ECOD & 0.978 &	0.962 &	0.969 \\
                \hline
                    KNN	& 0.977	& 0.975 & 0.976 \\
                \hline
                   OSVM & 0.974 & 0.86	& 0.913\\
                \hline
                   IForest & 0.971 & 0.986	& 0.979 \\
                \hline
                   USAD & 0.975 & 0.856 & 0.911 \\
                \hline

    \end{tabular}
            \caption{Results of each of the TSFs on their corresponding testing set having anomalous sliding windows from the real dataset.}
        \label{table:4}
\end{center}
\end{table}

\subsection{AD Model Selection Based on an RL Framework}

In this work, we proposed an AD model selection using a RL framework.  
The original reward function, shown in equation \ref{eq3}, was used in our proposed RLAD model selection framework. 
The performance of the proposed model selection framework for AD the synthetic dataset is stated in Table \ref{table:1}. Comparing the results of the six AD models in Table \ref{table:1}; shows that the proposed AD model selection provided a precision of 1, the same as KNN outperforming all the remaining AD models. In terms of the F1score, our suggested AD model selection recorded a value of 0.989 higher than the entire candidate AD models except for KNN having a slightly greater F1score of 1. 
In the case of the real dataset, results in  Table \ref{table:2} show that the proposed AD model selection, having a precision of 0.977 and an F1score of 0.727, outperformed all the candidate AD techniques of Table in terms of the precision and F1score. 

\subsection{Comparison of AD Model Selection Based on an RL Framework to an LLM-based AD model}
\begin{figure}[h]
            \centering
            \includegraphics[width=7cm]{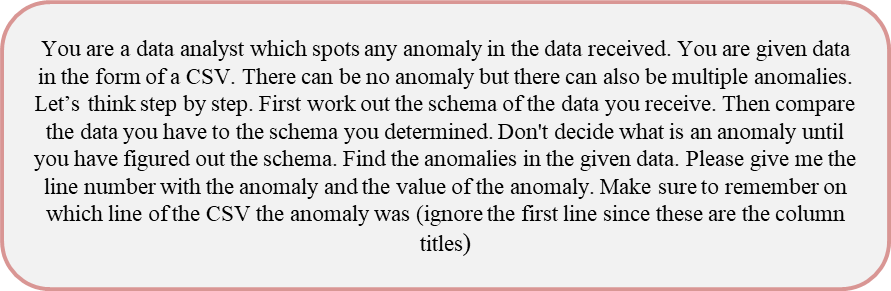}
            \caption{Template for the prompt strategy for the LLM to perform an AD task}
        \label{Figure:2a}              
\end{figure}

Considering the synthetic dataset, which is a relatively small dataset, the proposed RLAD model selection framework was compared to an LLM-based AD model. Leveraging Chain-of-Thought (CoT) prompting, as illustrated in Figure \ref{Figure:2a}, we motivated the ChatGPT4o-mini model to perform an AD task and to break down the complex thoughts or "processes" into smaller steps before providing a response on any detected anomalies. The LLM model yielded encouraging results on the AD task scoring an F1 score of 0.913 and a recall of 0.84 as shown in Table \ref{table:2a}. Yet, our proposed AD model framework, recording an F1 score of 0.989 and recall of 0.977 still outperforms the LLM-based AD model. 

\begin{table}[t]
\begin{center}       

    \begin{tabular}{ |c|c|c|c| } 
                \hline
                    \textbf{AD Model} & \textbf{Precision} & \textbf{Recall} & \textbf{F1score} \\ 

                \hline
                   LLM & 1 & 0.83 & 0.913 \\
                \hline
                    \textbf{Proposed} & \textbf{1} & \textbf{0.977} & \textbf{0.989}  \\
                \hline
    \end{tabular}
            \caption{Performance of the LLM-based AD model on the synthetic dataset in comparison of the proposed RLAD model selection framework.}
        \label{table:2a}
\end{center}
\end{table}

\subsection{Comparison of Different RL Frameworks for AD Model Selection}
The proposed RLAD model selection framework, which uses a small portion of ground truth labels, was compared with two other RLAD model selection frameworks; The first framework RLAD$_{Gtruth}$  totally relies on ground truth labels in its reward function. This means that, in all the instances, the reward was obtained by comparing the predicted anomaly label with the ground truth label. On the contrary, the second framework named RLAD$_{Class}$ relies totally on the predicted class labels obtained from TSFs. It should be noted that the RLAD$_{Class}$ framework also requires a portion of ground truth labels (used for TSFs training). The state-space and action-space remain the same in all frameworks. The original reward of equation \ref{eq3} was applied in all of the three frameworks (r$_{TN}$= 0.5, r$_{FN}$= -3, r$_{TP}$= 1, r$_{FP}$= -1.5).

The precision, recall, and F1score results of these two AD model selection frameworks based on RL are also present in Tables \ref{table:5} and \ref{table:6} using the synthetic and real datasets respectively. In both datasets, all the frameworks showed nearly the same high performance. In the synthetic data, the three frameworks recorded a precision of 1; however, our proposed framework recorded the highest F1score of 0.989.  In the real data, our proposed framework, having a 0.977 precision and an F1score of 0.727, showed a slightly higher performance than RLAD$_{Class}$ but slightly lower than RLAD$_{Gtruth}$. It can be said that, in the case of the real dataset, the proposed AD model selection offered a compromise between RLAD$_{Gtruth}$ that explicitly uses ground truth labels and RLAD$_{Class}$ that relies totally on class labels in its reward function.    

\begin{table}[t]
\begin{center}       

    \begin{tabular}{ |c|c|c|c| } 
                \hline
                    \multicolumn{4}{|c|} {\text{Synthetic Dataset}}\\
                \hline
                    \textbf{AD Model} &\textbf{Precision} & \textbf{Recall} & \textbf{F1score} \\                 
                \hline
                    \textbf{Proposed} & \textbf{1 } & \textbf{0.977} & \textbf{0.989} \\
                \hline
                    RLAD$_{Gtruth}$ & 1 & 0.955 & 0.977 \\
                \hline
                    RLAD$_{Class}$  & 1 & 0.925 & 0.961 \\
                \hline
    \end{tabular}
            \caption{Results of the three different RL Model Selection frameworks for AD on the anomalous synthetic data.}
        \label{table:5}
\end{center}
\end{table}

\begin{table}[t]
\begin{center}       

    \begin{tabular}{ |c|c|c|c| } 
                \hline
                 \multicolumn{4}{|c|} {\text{Real Dataset}}\\
                \hline
                    \textbf{AD Model} &\textbf{Precision} & \textbf{Recall} & \textbf{F1score} \\                 
                \hline
                    \textbf{Proposed} & 0.977 & 0.579 & 0.727 \\
                \hline
                    RLAD$_{Gtruth}$ & 0.971 & \textbf{0.67} & \textbf{0.793}  \\
                \hline
                    RLAD$_{Class}$  & \textbf{0.979} & 0.549	& 0.703 \\
                \hline
    \end{tabular}
   
            \caption{Results of the Performance of the three different RL Model Selection frameworks for AD on the anomalous real data.}
        \label{table:6}
\end{center}
\end{table}

\subsection{Different Constant Reward Functions with a Decaying Epsilon }
To study the effect of the different reward values on the proposed RL framework, we considered two extreme scenarios of reward functions R1 and R2 (Table \ref{table:7}) and evaluated them on the synthetic dataset. It can be observed that in comparison with the original reward function (r$_{TN}$= 0.5, r$_{FN}$= -3, r$_{TP}$= 1, r$_{FP}$= -1.5), the TN reward in R1 was decreased from 0.5 to 0.15. The FP reward in the original reward which represented a high penalty of -3 was changed into a low positive reward of 0.1. However, the FN reward and TP rewards in both functions remained the same (r$_{FN}$= -3 and r$_{TP}$= 1). Applying a lower positive reward on TN and a low positive reward on FP instead of a high penalty has led to a decrease in the model’s performance in the case of R1. In this regard, the precision decreased sharply from 1 to 0.636, the recall decreased from 0.977 to 0.88, and the F1score also decreased from 0.989 to 0.738 as shown in Table \ref{table:8}.

As shown in Table \ref{table:7}, another extreme scenario was investigated in the R2 function; in which the TN reward in R2 was set to be greater than the TN reward in the original function. Similarly, the FP penalty in R2 was set to be greater than the FP penalty in the original function. However, the penalty applied on FN in the original reward function was modified to a low reward in the case of R2. The TP reward in R2 was set to be less than the TP reward in the original function. The results of the original reward function in comparison to R2 reward function, in Table \ref{table:8}, reflected a slight decrease in the precision from 1 to 0.987 but a significant decrease in the recall from 0.977 to 0.564, and in the F1score from 0.989 to 0.718.  This is due to decreasing the TP reward and giving the agent a low reward in case of a FN instead of penalizing it with a negative reward.
As a result, the reward functions play a significant role in the performance of the proposed AD model selection approach in which the original reward function offered the most suitable and logical formulation of the AD model selection problem.

\subsection{Adaptive Rewards with Different Epsilons}
\begin{figure}[h]
            \centering
            \includegraphics[width=8cm]{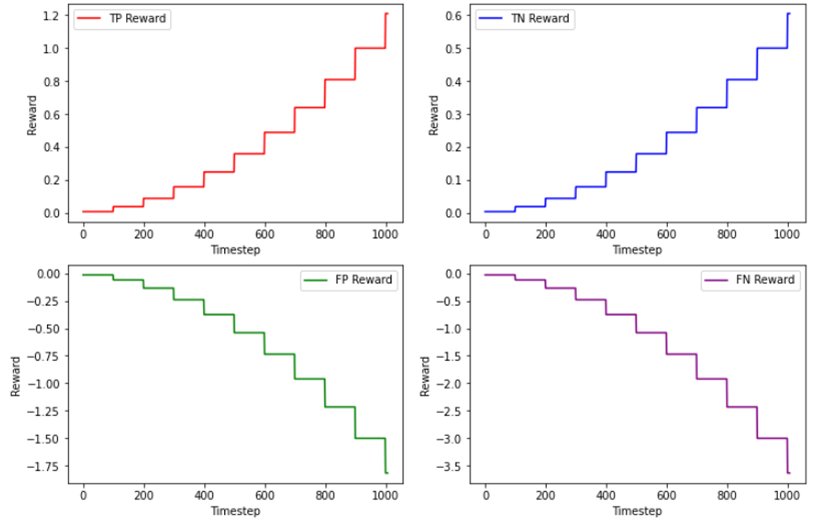}
            \caption{Adaptive Increasing Reward}
        \label{Figure:2}              
\end{figure}

We also inspected the effect of two adaptive reward functions; an increasing adaptive reward AdapInc and a decreasing adaptive reward AdapDec. 
In the AdapInc, we designed a reward function that increases iteratively with the time steps of the anomalous data. In this regard, the TN and TP rewards were intended to increase in a step-wise parabolic manner as a function of the time steps (or length) of the anomalous data. Similarly, the FN and FP penalties were set to rise in a step-wise parabolic form as a function of the time steps of the anomalous data. An illustration of the AdapInc reward is demonstrated in Figure \ref{Figure:2}. The AdapInc reward is formulated in the equation below:

\begin{equation}
R_{\text{AdapInc}} = \begin{cases}
    r_{\text{TN}} = 5 \times 10^{-2} \times C^2      & \text{TN} \\
    r_{\text{FN}} = -3 \times 10^{-2} \times C^2     & \text{FN} \\
    r_{\text{TP}} = 1 \times 10^{-2} \times C^2      & \text{TP} \\
    r_{\text{FP}} = -1.5 \times 10^{-2} \times C^2   & \text{FP}
\end{cases}
\end{equation}

In both adaptive rewards, the counter C was initially set to 1, for every 100 time-steps the counter would be incremented by 1. When reaching the final time-step (final sliding window of the anomalous dataset), C would be reset to 1.
Alternatively, as shown in Figure 4, the reward in the AdapDec is designed to decrease as a function of the time-steps of the anomalous data. Towards this, the TN and TP rewards would decrease iteratively step-wise hyperbolically. Also, the FN and FP penalties would decline in a step-wise hyperbolic manner as a function of the time steps (or length) of the anomalous data. The formulation of the AdapDec reward is stated in the equation below:

\begin{equation}
R_{\text{AdapDec}} = \left\{
\begin{array}{ll}
r_{TN} = \frac{0.5}{C^2}       & \text{TN} \\
r_{FN} = \frac{-3}{C^2}   & \text{FN}  \\
r_{TP} = \frac{1}{C^2}    & \text{TP}   \\
r_{FP} = \frac{-1.5}{C^2}  & \text{FP}
\end{array}
\right.
\end{equation}

\subsubsection{Adaptive Reward Functions with a Decaying Epsilon }
In this experiment, the same decaying exploration epsilon of Figure 2 was used. The proposed model selection AD framework was evaluated on the two adaptive rewards. The AdapInc reward, having an increasing reward function, encourages the agent to explore different actions at the beginning of the iterations (time-steps of the data) which gives the agent a higher chance to select randomly an AD model at the beginning. Then later on, on higher iterations, the agent is invited to greedily select a model that would yield the largest reward. Hence, eventually, the agent would be able to learn the optimal policy of selection. This is reflected in the results of the proposed AD model selection. The original and AdapInc rewards in Table 9 both had a precision of 1. The F1score of AdapInc reward is 0.981, slightly lower than that of the original reward with an F1score of 0.989.
On the other hand, the AdapDec reward, representing a decreasing reward function, prevents the agent from exploring various actions at the beginning and forces the agent at the beginning to select greedily an action (an AD model) that gives him a high reward which does not allow the agent to learn the optimal policy of selection. This is proven in the results of the proposed AD model selection in Table \ref{table:9} with AdapDec having a precision of 0.763 and an F1score of 0.833.
Given that the decaying exploration epsilon, which already allows the agent to explore more in the earlier time steps of the RL training phase, was applied in all cases, both the original and the AdapInc rewards achieved very similar high performance as shown in Table \ref{table:9}. On the contrary, the AdapDec, which prevents exploration in the earlier time instances, recorded a deteriorating performance. 

\subsubsection{Adaptive Reward Function with a Constant Exploration Epsilon }
Considering the proposed model selection framework, we also evaluated the same original reward, adaptive increasing reward AdapInc, and adaptive decreasing reward AdapDec but instead of using a decaying exploration epsilon, a constant exploration epsilon was applied. We considered three various cases; an exploration epsilon equals 0.9, 0.5, and 0.05.  As stated in Table \ref{table:9}, in case of a high epsilon value of 0.9, all three rewards showed high performance where the AdapInc reward recorded the highest evaluation metrics with an F1score of 0.974, followed by the original reward having an F1score of 0.933, and finally the AdapDec reward with an F1score of 0.925. Upon setting the epsilon to 0.5, the original reward and the AdapInc reward both revealed good performance in which the original reward showed the best results opposing that of an AdapDec reward which exhibited a degrading performance. The F1score of the original, AdapInc, and AdapDec rewards, as shown in Table 9, were 0.953, 0.938, and 0.743 respectively. In the case of a very low epsilon of 0.05, the original reward produced the highest F1score of 0.942, while the AdapDec reward yielded the lowest F1score of 0.826. As for the AdapInc reward, it resulted in an F1score of 0.92.
As a result in all three cases, both the original and the AdapInc rewards showed better results. However, the original reward had the highest score in the cases of 0.5 and 0.05. This implies that, in the case of low and intermediate exploration epsilons, a constant reward yielded the best performance. Whereas, in the case of a high exploration epsilon (epsilon= 0.9), the AdapInc reward gave the highest scores. 
All in all, the original reward and the AdapInc reward functions maintained both higher results in the different studied cases. However, the original reward using a decaying exploration epsilon produced the best overall performance.

\begin{figure}[h]
            \centering
            \includegraphics[width=8cm]{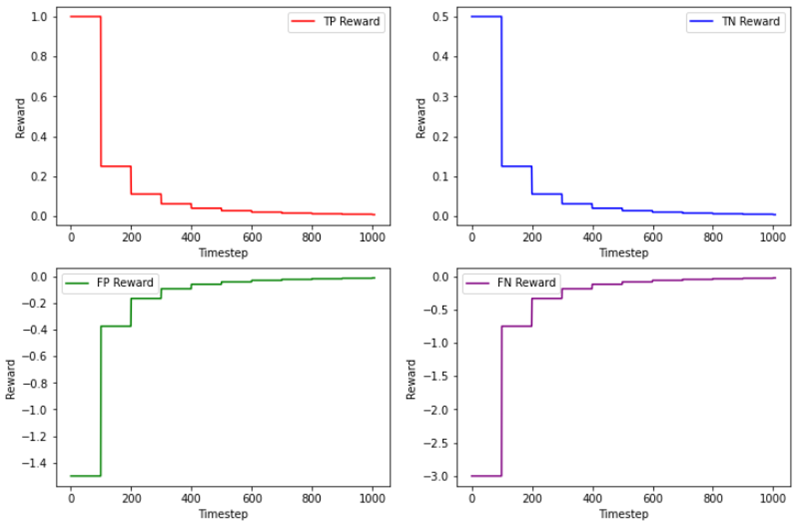}
            \caption{Adaptive Decreasing Reward}
        \label{Figure:3}            
\end{figure}

\begin{table}[t]
\begin{center} 
\begin{tabular}{|c|c|c|c|c|} 
                \hline
                 \multicolumn{5}{|c|}{\text{Reward Function}}\\        
                \hline
                    \textbf{Reward} & \textbf{TN} & \textbf{FN} & \textbf{TP} & \textbf{FP} \\                 
                \hline
                    \textbf{Original} & \textbf{0.5} & \textbf{-3} &	\textbf{1} & \textbf{-1.5} \\
                \hline
                    R1 & 0.15	& -3 & 1 & 0.1 \\
                \hline
                    R2 & 1 & 0.1 & 0.15 & -3 \\
                \hline
\end{tabular}

    \caption{Different Constant Reward Functions on the Proposed RL Framework for AD Model Selection.}
    \label{table:7}
\end{center}
\end{table}

\begin{table}[t]
\begin{center} 
\begin{tabular}{|c|c|c|c|} 
                \hline
                \multicolumn{4}{|c|}{\text{Decaying Epsilon}}\\            
                \hline
                    \textbf{Reward} & \textbf{Precision} & \textbf{Recall} & \textbf{F1score} \\                 
                \hline
                    \textbf{Original} & \textbf{1} & \textbf{0.977} &	\textbf{0.989} \\
                \hline
                    R1 & 0.636 & 0.88 & 0.738 \\
                \hline
                    R2 & 0.987 & 0.564 & 0.718 \\
                \hline
            \end{tabular}
            
    \caption{Results of the Performance of the Proposed RL Framework for AD Model Selection on Different Constant Reward Functions Using the Synthetic Dataset.}
    \label{table:8}
\end{center}
\end{table}

\begin{table*}[h]
\centering 
\begin{tabular}{|c|c|c|c|c|} 
\hline
\bfseries Reward Function & \bfseries Epsilon & \bfseries Precision & \bfseries Recall & \bfseries F1score \\
\hline
Original & & \textbf{1} & \textbf{0.977} & \textbf{0.989} \\
AdapInc & Decaying & 1 & 0.962 & 0.981 \\
AdapDec & & 0.763 & 0.917 & 0.833 \\
\hline
Original & & 0.983 & 0.887 & 0.933 \\
AdapInc & Constant=0.9 & 0.977 & 0.97 & 0.974 \\
AdapDec & & 0.975 & 0.88 & 0.925 \\
\hline
Original & & 1 & 0.91 & 0.953 \\
AdapInc & Constant=0.5 & 0.976 & 0.902 & 0.938 \\
AdapDec & & 0.675 & 0.827 & 0.743 \\
\hline
Original & & 0.968 & 0.917 & 0.942 \\
AdapInc & Constant=0.05 & 0.944 & 0.895 & 0.92 \\
AdapDec & & 0.784 & 0.872 & 0.826 \\
\hline
\end{tabular}
\caption{Results of the Performance of the Proposed RL Framework for AD Model Selection on Different Reward Functions Using the Synthetic Dataset.}
\label{table:9}
\end{table*}

\subsection{Different Types of Anomalies} 
In time series data, there exist different types of anomalies, consisting of global anomalies, local anomalies, and clustered anomalies.

\textbf{Global anomalies:} point anomalies that seem to differ from the overall time series. 

\textbf{Local anomalies:} point anomalies that deviate from their neighborhood data points. A point anomaly takes place at any time but does not repeat. 

\textbf{Clustered anomalies:} represent a group of anomalies that show similar characteristics. Clustered or sequential anomalies may be present once or repeatedly. Clustered anomalies can be local or global anomalies [57]. 

Usually, the available real-world datasets have a mixture of various anomaly types. This makes it difficult to recognize how AD techniques would behave in certain types of anomalies [16]. In this work, we attempted to use the time series of hourly electricity consumption data from the residential areas of France also provided by ENTSO-E Transparency [58]. The hourly power-consumption data from the year 2015 until 2017 having a total of 26302 time-steps is considered for the normal data. The remaining 8016 time-steps of the year 2018 will be used for the anomalous dataset.
To evaluate the effect of different anomaly types, we created three separate anomalous datasets; global, local, and clustered, anomalous datasets, in which we injected a quantity of 2\% anomalies to the hourly power-consumptions of 2018 of global, local, and clustered types respectively. 
We applied the same preprocessing procedure that was performed on the above synthetic dataset of Germany and then trained the six unsupervised AD techniques on the normal sliding windows and evaluated their performance on each of the three anomalous sliding windows. 
The precision, recall, and F1score of the AD models on each of the global, local, and clustered anomalous datasets in Tables \ref{table:13} shows that the performance of AD models differed depending on the type of anomalies. Except for KNN and IForest, all AD models showed low performance in both datasets with local and global anomalies and higher performance in the dataset with clustered anomalies. KNN and IForest, on the other hand, maintained a high performance in all cases recording the highest performance in the global dataset and the lowest in the clustered dataset.
To apply the proposed AD model selection on each of the three global, local, and clustered anomalous datasets, the same process of implementing six TSFs was performed to classify the predicted anomaly labels of each AD model into correct labels or wrong labels in an attempt to avoid using large quantities of ground truth labels. The results of the performance of each of the TSFs on the three different anomalous testing sets are provided in Tables \ref{table:10}, \ref{table:11}, and \ref{table:12} where each TSF showed high-performance scores. 
After that, using the anomalous sliding windows, the predicted anomaly scores from the AD models, the obtained empirical thresholds of the AD models, and their predicted anomaly labels, as well as the two confidence scores, the proposed AD model selection framework was applied for each of the three distinct anomalous datasets.
 Table 14 showed the precision, recall, and F1score on each of the three different types of anomalous datasets. The proposed model selection framework achieved an F1score of 0.962, 0.931, and 0.963 in the global, local, and clustered datasets respectively outperforming the performance of the whole AD models except for KNN. KNN yielded a slightly higher performance with F1score of 1, 0.99, and 0.965. The reason behind this is that in our proposed AD model selection framework, the reward function mostly relies on the predicted class labels of the TSF instead of explicitly depending on ground truth labels. And although the classifications of the TSFs are considered accurate, it is not 100\% accurate which would slightly decrease the performance of our proposed AD model selection in some cases. Nonetheless, in all the considered datasets, our proposed model selection approach for AD maintained high-performance results. 
\begin{table}[h]
\begin{center}       

    \begin{tabular}{|p{2.5cm}|c|c|c|c| } 
                \hline
                    \textbf{TSF based on predictions from} & \textbf{Precision} & \textbf{Recall} & \textbf{F1score} \\ 
                \hline
                    COPOD & 0.997 &	0.921 &	0.958 \\
                \hline
                    ECOD & 0.946 &	0.99 &	0.967 \\
                \hline
                    KNN	& -	& - & - \\
                \hline
                   OSVM & 0.999 &	0.999 &	0.999\\
                \hline
                   IForest & 0.999 & 0.996	& 0.998 \\
                \hline
                   USAD & 0.987 &	0.918 &	0.951 \\
                \hline

    \end{tabular}
            \caption{Results of each of the TSFs on their corresponding testing set having global anomalous sliding windows from the France dataset.}
        \label{table:10}
\end{center}
\end{table}

\begin{table}[h]
\begin{center}       

    \begin{tabular}{|p{2.5cm}|c|c|c|c| } 
                \hline
                    \textbf{TSF based on predictions from} & \textbf{Precision} & \textbf{Recall} & \textbf{F1score} \\ 
                \hline
                    COPOD & 0.991 &	0.992 &	0.991 \\
                \hline
                    ECOD & 0.932 &	0.999 &	0.964 \\
                \hline
                    KNN	& 0.999 &	1 &	0.999 \\
                \hline
                   OSVM & 0.998 &	0.999 &	0.999\\
                \hline
                   IForest & 0.996 &	0.988 &	0.992 \\
                \hline
                   USAD & 0.998 &	0.981 & 0.989 \\
                \hline

    \end{tabular}
            \caption{Results of each of the TSFs on their corresponding testing set having local anomalous sliding windows from the France dataset.}
        \label{table:11}
\end{center}
\end{table}

\begin{table}[h]
\begin{center}       

    \begin{tabular}{|p{2.5cm}|c|c|c|c| } 
                \hline
                    \textbf{TSF based on predictions from} & \textbf{Precision} & \textbf{Recall} & \textbf{F1score} \\ 
                \hline
                    COPOD & 0.997 & 0.999 &	 0.998 \\
                \hline
                    ECOD &  0.991 &	1	& 0.995 \\
                \hline
                    KNN	&  0.999 &	1 &	0.999 \\
                \hline
                   OSVM &  0.998	&1 &	0.999\\
                \hline
                   IForest & 0.998	& 0.999	& 0.999 \\
                \hline
                   USAD & 0.998 &	1 &	0.999 \\
                \hline

    \end{tabular}
            \caption{Results of each of the TSFs on their corresponding testing set having clustered anomalous sliding windows from the France dataset.}
        \label{table:12}
\end{center}
\end{table}

\begin{table}[h]
\begin{center} 
\begin{tabular}{|p{2.5cm}|c|c|c|c|c|}
                \hline
                    \textbf{F1score of AD Model} & \textbf{Global} & \textbf{Local}  & \textbf{Clustered} \\ 
                \hline
                    COPOD & 0.283 & 0.088 & 0.611 \\
                \hline
                    ECOD & 0.241 & 0.117 & 0.594 \\
                \hline
                    KNN	& 1 & 0.99	& 0.965 \\
                \hline
                   OSVM & 0.093 & 0.016 & 0.611 \\
                \hline
                    IForest & 0.922	& 0.865 & 0.852 \\
                \hline
                   USAD & 0.286 & 0.105 & 0.747 \\
                \hline
                    \textbf{Proposed} & \textbf{0.962} & \textbf{0.931} & \textbf{0.963}  \\ 
                \hline
\end{tabular} 
    \caption{Results of the AD Models on Different Anomaly Types.}
    \label{table:13}
\end{center}
\end{table}

\section{Conclusion and Future Work}
In this study, we address the challenges posed by the complex and diverse nature of data abnormalities in real-world systems, coupled with the limited availability of large quantities of ground truth labels. To tackle this, we propose a model selection framework for AD that combines RL with TSF. Our approach incorporates partial reliance on ground truth labels in the reward function and dynamically selects an appropriate AD model at each time step.

The results of our proposed approach are remarkable. It surpasses all other AD models in terms of performance on real-world data and achieves outstanding results on synthetic data, outperforming almost all other models except for KNN, which attains an F1 score of 1. It also exceeded the performance the LLM-based AD model on the synthetic dataset. 

The proposed AD model selection approach was examined using different reward functions. The original and AdapInc rewards functions exhibited superior results in both constant and decaying exploration epsilon scenarios. The original reward function with a decaying exploration epsilon demonstrated the best precision, recall, and F1 score outcomes.

The impact of different types of anomalies was investigated, revealing variations in the performance of each AD model. This finding further emphasizes the significance of the proposed model selection framework for unsupervised AD, enabling accurate detection of anomalies in time series data without an extensive availability of ground truth labels.

Future endeavors will involve applying the proposed AD model selection framework to larger time series datasets with higher dimensions. Additionally, incorporating additional semi-supervised AD models into the pool of candidate models is an avenue for further exploration. In this regard, we plan to incorporate additional semi-supervised AD techniques. Given the promising results of LLMs in AD tasks, we are also considering the integration of an LLM-based AD model into the pool of candidate models. By requiring only a small portion of ground truth labels and harnessing the strengths of multiple AD models, our approach raises its applicability to real-world scenarios with diverse anomaly types. 

\section{Acknowledgement}
\label{Acknowledgement}
This work was supported by the Maroun Semaan Faculty of Engineering and Architecture and the University Research Board at the American University of Beirut, Lebanon.



\appendix

\section{Hyperparameter Choices}
\label{sec:sample:appendix}

\begin{table}[h]
\begin{tabular}{|p{4cm}|p{4cm}|p{4cm}|c|}
    \hline
    \textbf{Hyperparameter 1} & \textbf{Hyperparameter 2}  \\ 
    \hline
    n\_neighbors: [1, 5, 10, 15, 20, 25, 50, 60, 70, 80, 90, 100] & method: ['largest', 'mean', 'median'] \\
    \hline
    nu (train error tol): [0.1, 0.2, 0.3, 0.4, 0.5, 0.6, 0.7, 0.8, 0.9] & - \\
    \hline
    n\_estimators: [10, 20, 30, 40, 50, 75, 100, 150, 200] & max\_features: [0.1, 0.2, 0.3, 0.4, 0.5, 0.6, 0.7, 0.8, 0.9]\\
    \hline
\end{tabular}
\setcounter{table}{0}
\caption{Hyperparameter Choices.}
\label{tab:hyperparameters}
\end{table}
\end{document}